\documentclass[lettersize, journal, doublecolumn]{IEEEtran}
\usepackage[english]{babel}

\usepackage{subcaption}
\DeclareCaptionLabelSeparator{periodspace}{.\quad}
\usepackage[]{footmisc}

\usepackage{cite}
\usepackage{amsmath,amssymb,amsfonts}
\usepackage{algorithmic}
\usepackage{graphicx}
\usepackage{textcomp}
\usepackage{xcolor}
\usepackage{float}
\usepackage{url}
\usepackage{rotating}
\usepackage{gensymb}
\PassOptionsToPackage{hyphens}{url}\usepackage{hyperref}

\def\BibTeX{{\rm B\kern-.05em{\sc i\kern-.025em b}\kern-.08em
    T\kern-.1667em\lower.7ex\hbox{E}\kern-.125emX}}

\makeatletter
\newcommand{\linebreakand}{
  \end{@IEEEauthorhalign}
  \hfill\mbox{}\par
  \mbox{}\hfill\begin{@IEEEauthorhalign}
}
\makeatother

\title{Neural Networks for Path Planning}

\author{\IEEEauthorblockN{Salim Janji},~\IEEEmembership{Graduate Student Member,~IEEE}, \IEEEauthorblockN{Adrian Kliks},~\IEEEmembership{Senior Member,~IEEE}}

\begin{document}
\maketitle

\begin{abstract}
The scientific community is able to present a new set of solutions to practical problems that substantially improve the performance of modern technology in terms of efficiency and speed of computation due to the advancement in neural networks architectures. We present the latest works considering the utilization of neural networks in robot path planning. Our survey shows the contrast between different formulations of the problems that consider different inputs, outputs, and environments and how different neural networks architectures are able to provide solutions to all of the presented problems.
\end{abstract}

\section{Introduction}
Robot path planning has been receiving increased interest within the research community due to the numerous emerging technologies it advances \cite{survHeuristic}. The most prominent of these technologies include robots \cite{robotPlanning}, autonomous vehicles \cite{surveyVehicles} and UAV path planning \cite{surveyUAV}. The problem of path planning is concerned with finding an optimal path between the source and destination. An optimal decision usually takes into consideration the following aspects:
\begin{itemize}
    \item Minimizing traversed distance.
    \item Reducing the magnitude of turning angles.
    \item Avoiding both static and dynamic obstacles with a certain clearance.
\end{itemize}
Planning the path completely requires consideration of two layers of planning: The first layer develops a low-resolution high-level solution which is the goal of an \textit{offline path planner} or \textit{global path planning}. This is usually performed using an existing environment map which clearly does not take into consideration dynamic obstacles or unexpected events. To deal with the aforementioned dynamicity an \textit{online} or \textit{local path planner} is required which results in a high-resolution low-level path that guides the robot locally in a slice along the previously determined global path.

Depending on the space dimensionality, number of sensors acting as inputs, dynamicity of the environment, and different movement constraints such as maximum speed and turning angles, the problem of finding an optimal path can become increasingly complex \cite{complexDynamics}. To cope with the increased complexity of path planning modules, and to reduce the computation time required during operation, neural networks (NN) have been employed in various path planning solutions \cite{surveyNeuralPath, generalizedNeuralNetwork, neuralRrt, neuralRrt3d,vins, onTheTraining, segmentizedView, neuralDynamics2020}. In this paper we present the most recent works that utilize NNs in solving the robot path planning problem.

The paper is structured as follows. In section \ref{NRRT2D} we present the scheme proposed in \cite{neuralRrt} for 2D path planning which is based on rapidly random-exploring tree (RRT), followed by its generalization to 3D problem by the same authors which we present in section \ref{nrrt3d} based on \cite{neuralRrt3d}. Then, in section \ref{sec:vins}, a framework utilizing value iteration networks (VINs) for UAV 3D online path planning is described based on the work in \cite{vins}. The authors in \cite{onTheTraining} demonstrated the differences in performance when a multi-layer perceptron (MLP) network is trained using different offline-path planning algorithms and with different environment. We present their work and results in section \ref{sec:mlp}. In \cite{segmentizedView}, the authors consider the existence of moving obstacles and devise a scheme for online path planning which we present in \ref{sec:segmentized}. Finally, in \cite{neuralDynamics2020} the authors present the weaknesses in utilizing neural dynamics \cite{neuralDynamicsOld} to find paths, and propose modifications to overcome them. We present their findings and proposal in section \ref{sec:neuraldynamics} and then conclude our paper with section \ref{sec:conclusion}.

\section{Neural Rapidly Random-exploring Tree* (NRRT*)}\label{NRRT2D}
\subsection{RRT and RRT*}
RRT was first introduced in \cite{rrt} to design high-dimensional spaces by randomly building a space-filling tree. In the case of robot path planning, and when using uniform sampling distributions, such sampling methods like RRT and probabilistic roadmap (RPM) provide \textit{probabilistic completeness} in the sense that by allowing the sampling process to run infinitely, the probability of sampling all the points in the state space increases monotonically. Furthermore, they ensure \textit{asymptotic optimality} which means that the algorithm is guaranteed to converge to the optimal solution asymptotically. 

\begin{figure}[!htp]
\centering
\centerline{\includegraphics[width=0.3\textwidth]{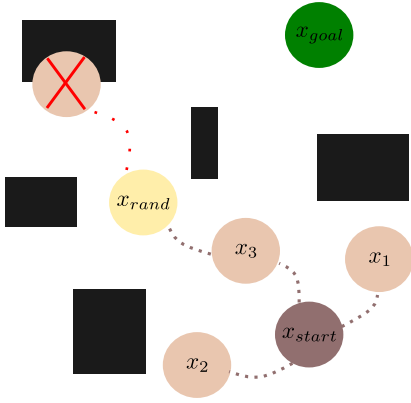}}
\caption{RRT sampling.}
\label{fig:rrt_sampling}
\end{figure}

Figure \ref{fig:rrt_sampling} demonstrates the process of RRT in the context of path planning. Specifically, the process begins with the starting node, $x_{\text{start}}$, where different points in space are sampled around it. If a sampled point, $x_{\text{rand}}$, does not overlap with any obstacle, then that sample is added to the tree as a leaf node to the nearest parent ($x_3$ in the figure). In the enhanced version RRT*, the procedure further includes a \textit{ChooseParent} and \textit{Rewire} processes. The former selects the parent for the new leaf node based on a defined cost minimization, and the latter process will test neighbor vertices to check if a path with less cost is now available through the new node. 

Clearly, if a solution exists, then RRT* is guaranteed to find it. However, in real-life implementations, this process consumes a lot of time and memory until it reaches the optimal solution which motivated the authors in \cite{neuralRrt} to consider the utilization of NNs to speed-up the convergence such that the proposed mechanism can be used as an online-planning scheme. 

\subsection{NRRT*}
Instead of uniformly sampling the sate space, the authors in \cite{neuralRrt} propose using a convolutional neural networks (CNNs) architecture, ResNet50 \cite{ResNet50}, to generate the sampling distribution of the state space.

\subsubsection*{Input} Specifically, the NN architecture receives as inputs the following:
\begin{enumerate}
    \item The map of the environment whereas pixels are encoded according to
    \begin{equation}
    p\in
    \begin{cases}
        0 & \text{empty,}\\
        1 & \text{obstacle,}\\
        3 & \text{start position, }\\
        4 & \text{goal position.}\\
    \end{cases}
    \end{equation}
    \item The required step size, and clearance values. 
\end{enumerate}

\subsubsection*{output} The output of the proposed neural architecture is a map with the same dimensions of the input map such that each pixels contains a value $\hat{p}\in[0,1]$ which denotes the probability of this pixel being chosen during the sampling process. To preserve the asymptotic optimality and probabilstic completeness characteristics of RRT*, the authors propose sampling with equal probabilities from two sources. The first being the results of the NNs, and the second is the standard uniform sampling distribution which considers the whole space.

\subsubsection*{Architecture} the map is fed into a convolutional layer which extracts a high level feature map, which is in turn fed into another convolutional layer to extract low level feature maps with smaller size. On the other path related to the step size and clearance, the result of the first fully connected layer is concatenated with the low-level feature map, and the result of the second layer connected to the first one is concatenated with the high-level feature map. The map which contains the previously mentioned low-level feature map is passed through an Atrous convolution layer which is a form of dilated convolution. Convolving with differnt dilution rate allows us to generate maps with information based on different scales. The result is then concatenated to the result from high-level features path which contains a Atrous spatial pyramid pooling (ASPP) layer followed by upsampling to match the size of the other path. Finally, the resulting map is fed to a decoding convolutional layer which outputs the probability distribution map.

\subsubsection*{Training} To train the NNs architecture, the authors use a number of maps with optimal solutions generated using the A* algorithm \cite{a*}. The cost function for generating the optimal path penalizes both the traversed distance and turning angle. After obtaining the optimal path using A*, a widened version of the path is encoded in pixels on the map which is used for training.

\subsubsection*{Results} The authors simulate their mechanism for different input maps with different clearance and step-size parameters and prove that utilizing NRRT* is indeed able to speed up the convergence towards the optimal solution substantially.

\section{NRRT* 3D}\label{nrrt3d}
The same authors of the mechanism introduced in the previous section have also presented another NRRT* algorithm but for the case where the space is 3-dimensional \cite{neuralRrt3d}. To tackle this problem, they propose using one of two architectures: the first is based on U-Net \cite{u-net} which also makes use of CNNs, and the second utilizes a conditional generative adversarial network (cGAN).

\subsubsection*{Input} Unlike the 2D case, in the proposed solution the input only includes two 3D maps, one encoding the obstacles, and the second encoding the start and goal locations. That is, the clearance and step size are no longer considered as inputs.

\subsubsection*{output} The output is again a 3D distribution of the sampling space whereas favoured pixels are given a binary value of one, while the rest are encoded by a zero. The authors suggest that the algorithm is run in two phases. The first phase favors sampling from the NN output with a probability of 0.9 vs. 0.1 for sampling uniformly in the 3D space. After finding a solution, the second phase begins where the solution is to be improved by sampling uniformly (i.e., equal probability) from both sampling sources. Similarly to before, this ensures the \textit{probabilistic completeness} and the \textit{asymptotic optimality} of the mechanism.

\subsubsection*{Architecture} In the case of U-Net, the network is composed on encoding and decoding layers. Each map is first passed through a different convolutional layer, and the results are then concatenated. Then 4 3x3x3 convolutional layers are applied. In the decoding stage that follows, 4 3x3x3 deconvolutions are performed. A channel is defined as the combination of two layers, one from the encoding stage and the other with the same size from the corresponding decoding stage. Unlike U-Net, the authors propose adding the results of the maps in each channel before passing them to the next decoding layer.

As for the case of GAN, two networks are required, a generator and a discriminator. The goal of the generator is to fool the discriminator by generating noisy output. On the other hand, the discriminator is trained to distinguish correct output from the generator. By adding conditional information a cGAN is able to generate data that satisfies the start and end location conditions. Briefly, the generator network proposed includes a third map which contains the 3D noise. All three maps are passed through 3 seperate convolutional layers, then concatenation layer, and then passed through an network similar to the one explained above. The results is a probability distribution map which is fed to the discriminator network next to the environment and locations maps. The three maps are then passed through three separate convolutional layers and the results are concatenated. Finally, 4 successive convolutional layers with reduced kernel size result in a binary value which translates to a decision of \textit{fake} or \textit{real}.

\subsubsection*{Training} Similarly to the previous section, the optimal solution for each training map is obtained using A* algorithm, and then the NNs are trained using a diluted version of these paths. 

\subsubsection*{Results} The authors simulations prove that using NRRT* reduces the required number of sampling iterations by 20-30 times. Furthermore, they show that both NNs architecture are similar in performance, however, the cGAN model is able to improve slightly the accuracy and generalization ability.

\section{Online-planning with Value Iteration Networks (VINs)}\label{sec:vins}
The authors in \cite{vins} target the problem of UAV navigation by proposing a deep neural network framework which leverages VINs. They also test their mechanism with a UAV that is equipped with a ZED camera in both indoor and outdoor environment. They show impressive results in which the UAV is successfully able to navigate through obstacles towards a given goal location. The problem is formulated as a classification task with 26 possible actions (3 possible actions in each dimension corresponding to moving forward, backward, and staying still). 

\subsubsection*{Input} The inputs for the UAV are obtained through processing the camera output such that two 3D maps are built: a local 3D map encoding the obstacles within the specified range, and another map which encodes the agent and goal locations. In the second map, after adding the goal position, and before encoding the current position, a 3D convolutional layer is applied to spread the goal information to all pixels.

\subsubsection*{Architecture} The proposed framework is composed of three modules. We briefly describe each module:
\begin{enumerate}
    \item Decomposition module: this module slices the input maps along each 2D plane, with slices obtained around the local poisition in each map. Then a permutation of maps formed from the slices of each map and through all combinations of dimensions is obtained. Since only local information in each dimension is required, these slices are assumed to be sufficient for making the next decision.
    \item Planning module: for each map pair, the reward is first approximated for each possible action using two convolutional layers. Then, the results are fed into a VIN network for K iterations. The results of this module are Q values approximations for each map pair. In terms of reinforcement learning, the Q value corresponding to each action indicates how profitable it is.
    \item Composition module: this module concatenates the slices across plane and then concatenates the results again. Finally, the results are passed through convolutional layer, and then a fully connected layer which output the resulting Q value for each possible action. 
\end{enumerate}
The action corresponding to the largest Q value in the final output layer is then chosen. The authors propose using two instances of the proposed framework with different parameters (i.e., size of the local map) such that if the first preferred network fails to generate a collision free path, the output from the others can be selected.

\subsubsection*{Training} using Dijkstra’s algorithm in virtual environments, the authors trained their frameworks through imitation learning.

\subsubsection*{Results} First, it was shown that the computational time required per prediction is around 3 ms, then, through grid-world simulations, the proposed algorithm was shown to achieve superior performance in terms of success rate when compared to RRT offline planning. Also, using virutal environments simulation, the authors showed that their algorithm resulted in shorter travelling distance when compared to RRT*. Finally, the authors tested their mechanism with a real UAV and showed the applicability of their work to both outdoor and indoor environments.

\section{Training a Multilayer Perceptron (MLP) for Online Path Planning}\label{sec:mlp}

To deal with the uncertainties during navigation within a given environment in a timely manner, a NN can be employed such that it will determine the local path of a vehicle that is following a general path that is previously obtained with an offline planner. That is, the NN acts as an online planner. In contrast to the previously existing literature which describes how such a planner can be built and trained, the authors in \cite{onTheTraining} consider how different training data can affect the performance of such planners. 

\subsubsection*{NN Architecture} The authors utilize a six-layered NN with 50 neurons in each of the four hidden layers. The input layer consists of 3 neurons corresponding to the input, and the output layer is made of a single neuron. Finally, they use the recitified linear unit (ReLU) as the activation function.

\subsubsection*{Input} The vehicle state is represented by three parameters: $d_{obs}$, $a_{obs}$, and $a_{g}$ that denote the distance to the nearest obstacle, bearing angle to the nearest obstacle, and the bearing angle to goal direction, respectively.

\subsubsection*{Output} Given the state of the vehicle, $\Psi=\{d_{obs},\theta_{obs},\theta_{g}\}$, the NN outputs a value $\theta \in \{-\pi, \pi\}$ which serves as the steering angle deviating from the current direction.

\subsubsection*{Training} Similarly to the previous works, the authors train the NN with offline planning algorithms. They use two different algorithms and compare the achieved results to analyze the impact of algorithm selection. The first algorithm used is the Bellman-Ford (BF) algorithm which is commonly used to find the shortest path in a graph. Using this algorithm results in a discretized solution space obtained through decomposing the area into a grid whereas the vehicle can travel on 8 different paths from each vertex (i.e., through the square edges and diameters). 
In contrast, the second algorithm used implements a quadratic program (QP) which solves the optimization problem that minimizes the distance between the current position and goal position at each time instance given the obstacle clearance, maximum velocity, and maximum acceleration constraints.
For each case, the NN is trained using the optimal paths outputs resulting from 10,000 different problems, each with 10 obstacles in the map. 

\subsubsection*{Results} Then, the NNs were tested against two different sets of problems: the first one contained 30 maps with similar area and number of obstacles as the training set, and the second set contained 30 maps with double the area and with 80 obstacles instead of 10. It was found that using the BF algorithm with discrete solution space outperforms the QP scenario in both case. Specifically, the BF was able to succeed in avoiding all obstacles and finding the goal location in the first set of problem, whereas it failed at 5 out of 30 in the second more dense set. However, in the case of QP, the NN failed in 5 test maps of the first set, and 21 of the second set. An interesting observation is that when training the NN while using the BF algorithm, the NN was actually able to find paths with shorter distance than those obtained by the BF itself applied as an offline algorithm to the same maps. This is because the offline BF solution space is constrained to the grid waypoints, unlike the steering resulting from the NN.

\section{Path Planning with Moving Obstacles}\label{sec:segmentized}
\begin{figure}[!htp]
\centering
\centerline{\includegraphics[width=0.3\textwidth]{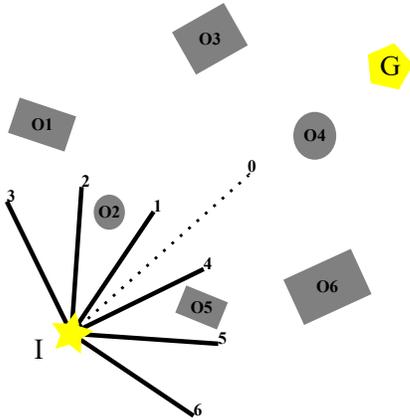}}
\caption{Navigating through moving obstacles.}
\label{fig:segmentized}
\end{figure}

The uncertainties in an environment can include moving obstacles as well, which was not considered by any of the previous works. However, the authors in \cite{segmentizedView} take into consideration the existence of obstacles moving with a constant speed less than a specified value in a random direction. As shown in figure \ref{fig:segmentized}, and at any time instant, the robot's field of view with an range of $150 ^{\circ}$ is divided into 5 segments (i.e., each segment contains $30 ^{\circ}$ of range) that are spread around the perpendicular line in the direction of the goal. 

The authors proceed by obtaining for each segment a \textit{critical obstacle} which is defined as the moving obstacle reaching the segment in the nearest time. That is, referring to figure \ref{fig:segmentized}, and by assuming that \textbf{O1} is a moving obstacle moving in the direction of \textbf{I}, then this obstacle is selected as the critical obstacle for segment \textbf{3I2}. Also, it is assumed that after each decision interval, the robot moves with a fixed distance of 20 units with a variable speed determined by the NN as will be explained below.

\subsubsection*{NN Architecture} The NN used is a 3-layer MLP with 25, 5, 5 neurons in the hidden, input, and output layers, respectively. As for the activation function, the authors use the sigmoid function.

\subsubsection*{Input} Each of the input layer neurons correspond to a different segment whereas the value of the input is as follows:
\begin{enumerate}
    \item If the segment is already occupied by an obstacle, then the input is 0
    \item Otherwise, the estimated time duration required for the critical obstacle to reach the segment is fed as an input.
    \item A value of 1.5 is fed if there are no obstacles blocking the segment, or critical obstacles approaching it.
\end{enumerate}

\subsubsection*{Output} For each segment, the NN outputs the following:
\begin{enumerate}
    \item 0 means no possible movement in that direction because it is blocked.
    \item A value ranging from 0.1 to 0.5 mean that the critical obstacle disallows movement in the corresponding segment.
    \item A value greater than 0.5 and less than 1.0 is interpreted as a decision for the robot to reach the destination within 0.5 seconds.
    \item Whereas if the output is greater than 1.0, then this means that the robot can head to the corresponding location within a duration of 1s.
\end{enumerate}
If more than one segment is allowed, the robot chooses the one that moves it closer to the goal.

\subsubsection*{Training} The authors train the network with given environments by manually calculating the desired outputs for each state, and performing backpropagation to fix the weights of the network.

\subsubsection*{Results} The authors show that the robot is able to avoid obstacles successfully for two different maps, and they compare the performance of their mechanism to other algorithms mentioned in literature in terms of computational time. They show that unlike other algorithms which require a computational time of at least 190 ms, forward propagation through the NN takes only 50 ms to output a response.

\section{Path-planning Based on Neural Activity Dynamics}\label{sec:neuraldynamics}
The authors in \cite{neuralDynamicsOld} first presented a path-planning solution based on a different type of neural networks. These neural networks provide a nerual activity map. We briefly present the solution below.

\subsubsection*{Input} The input needed for the mechanism described below is a map of the complete environment with obstacles locations marked along the goal and start positions.

\subsubsection*{NN Architecture} The neural map generated from the environment map contains a grid of neurons where each neuron is characterized by a shunting equation \cite{shunting} which determines the neuron's activity dynamics. In this model, each neuron is connected to its neighbouring neurons, and both excitatory and inhibitory signals flow throughout the network through these connections. The activity of each neuron, $x_i$, is obtained through solving the following shunting differential equation 

\begin{equation}
    \frac{dx_i}{dt} = - A x_i+(B-x_i) \underbrace{\left([I_i]^+ +\sum^N_{j=1}w_{ij}[x_j]^+\right)}_{S^+_i} - (D+x_i)[I_i]^-
\end{equation}

where $w_{ij}$ is the weight of the connection between the $j-th$ and $i-th$  and it is proportional to the distance between them up to a certain value where futher neurons are excluded, $[I_i]^+$ and $[I_i]^-$ denote the excitatory and inhibitory inputs respectively, and finally, $A$, $B$, and $D$ represent the passive decay rate, upper bound, and lower bound of the neural activity, respectively. The summation between excitatory input and positive neighbouring neuron activity, $[x_j]^+$, is denoted by $S^+_i$. To model the path planning problem, the inputs at a given neuron (i.e., location) are defined as follows:

\begin{equation}
 I_i=
\begin{cases}
    E, & \text{if there is a target} \\
    -E, & \text{if there is an obstacle}\\
    0, & \text{otherwise}
\end{cases}
\end{equation}
whereas $E>>B$ is a large positive constant. 

\subsubsection*{Output} Finally, after obtaining complete neural activity map values, and given the starting location, the robot selects the next neuron (i.e., spatial step) that has the maximum neural activity value taking into consideration an additional cost related to the turning angles of the robot. Therefore, the target \textit{attracts} the robot towards it through the propagation of excitatory input, while obstacles \textit{repel} the robot away.

\begin{figure}[!htp]
\centering
\centerline{\includegraphics[width=0.2\textwidth]{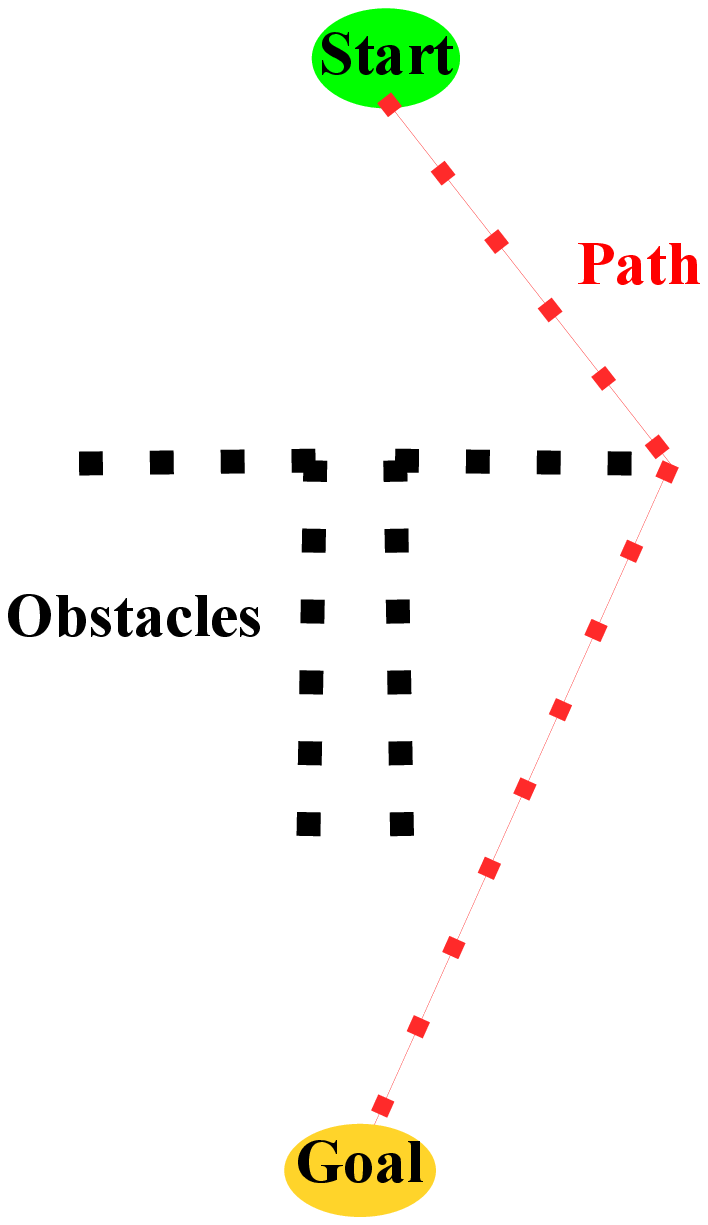}}
\caption{Sub-optimal path generated by neural dynamics.}
\label{fig:dynamic}
\end{figure}

However, the authors of the recent work \cite{neuralDynamics2020} proposed a modification to the proposed mechanism because it was found that the paths generated by this approach were sub-optimal in given situations. A plot of such a case which the authors presented is shown in figure \ref{fig:dynamic}

To fix this problem of sub-optimal paths generation in challenging environments, the authors propose two modifications that include padding and averaging operations. The padding operation pads neurons to the boundary of the map such that the activity value of each neuron is equal to the average of the neighbouring neurons inside the map. Secondly, they modified the excitatory input modeling $S^+_i$ as follows:

\begin{equation}
    S^+_i = [I_i]^+ + \frac{\left(\sum^N_{j=1}w_{ij}[x_j]^+\right)n-n_{\text{neg}}}{n_{\text{pos}}n}
\end{equation}

whereas $n_{\text{neg}}$ and $n_{\text{pos}}$ denote the number of neighbouring neurons with negative and positive activity, respectively, and $n$ is the total number of neighbouring neurons which also includes $n_\text{out}$ which is the number of neurons outside the map that are padded to the boundary of the map as explained above.

\subsubsection*{Results} The authors test their proposed algorithm in both dynamic (i.e., with moving obstacles) and static environments, and they compare their results with several other common path-planning solutions including those based on neural dynamics and neuro-fuzzy tools. The show cases where their proposed method was the only one able to find the optimal path with reduced number of turns and total turning angles. However, the clearance distance from obstacles resulting from their work is less on average from other solutions. 

\section{Conclusions}\label{sec:conclusion}
As presented, the application of neural networks to path-planning problems provides a substantial degree of freedom in terms of the problems inputs, outputs, and environments. Different inputs can exist in practical situations (i.e., sensors, cameras, maps, etc.), and neural networks can be designed appropriately as presented above. Furthermore, the output can take several forms such as turning angle, speed, next step location, etc. Finally, the neural network architecture can be designed depending on the amount of information that is available about the environment such as the speed of obstacles, map, or camera input, etc.

\bibliographystyle{ieeetr}
\bibliography{refs}
\end{document}